\documentclass[runningheads]{llncs}
\usepackage{graphicx}
\usepackage{amsmath,amssymb} 
\usepackage{color}
\usepackage{booktabs}
\usepackage[pagebackref,breaklinks,colorlinks]{hyperref}
\usepackage{xcolor}
\usepackage[algo2e, noline]{algorithm2e} 
\usepackage{bm}
\usepackage{multirow}
\usepackage{pifont}
\usepackage{subcaption}
\usepackage{algpseudocode}
\usepackage{bbding}
\usepackage[title]{appendix}
\usepackage{diagbox}
\usepackage{makecell}
\usepackage{colortbl}
\usepackage{wrapfig}

\RequirePackage{times}    
\RequirePackage{cite}     
\RequirePackage{xspace}
\makeatletter
\DeclareRobustCommand\onedot{\futurelet\@let@token\@onedot}
\def\@onedot{\ifx\@let@token.\else.\null\fi\xspace}
\def\@fnsymbol#1{\ensuremath{\ifcase#1\or *\or \dagger\or \ddagger\or
   \mathsection\or \mathparagraph\or \|\or **\or \dagger\dagger
   \or \ddagger\ddagger \else\@ctrerr\fi}}

\def\eg{\emph{e.g}\onedot} 
\def\ie{\emph{i.e}\onedot} 
 
\def\etc{\emph{etc}\onedot} 
\def\wrt{w.r.t\onedot} 
 
\def\etal{\emph{et al}\onedot}
\makeatother
\usepackage[capitalize]{cleveref}
\crefname{section}{Sec.}{Secs.}
\Crefname{section}{Section}{Sections}
\crefname{table}{Tab.}{Tabs.}
\Crefname{table}{Table}{Tables}
\crefname{figure}{Fig.}{Figs.}
\Crefname{figure}{Figure}{Figures}
\crefname{equation}{Eq.}{Eqs.}
\Crefname{equation}{Equation}{Equations}

\newcommand*{\mathcolor}{}
\def\mathcolor#1#{\mathcoloraux{#1}}
\newcommand*{\mathcoloraux}[3]{%
  \protect\leavevmode
  \begingroup
    \color#1{#2}#3%
  \endgroup
}
\definecolor{darkgreen}{RGB}{0,112,0}
\definecolor{mygray}{gray}{.9}
\begin{document}
\pagestyle{headings}
\mainmatter


\title{DSPNet: Towards Slimmable Pretrained Networks based on Discriminative Self-supervised Learning}
\titlerunning{DSPNet}
\authorrunning{S. Wang et al.}
\author{%
  Shaoru Wang\textsuperscript{\rm 1,3}\thanks{The work was done when Shaoru Wang was an intern in Megvii Inc.} \and Zeming Li\textsuperscript{\rm 2} \and Jin Gao\textsuperscript{\rm 1,3}\thanks{Corresponding author.} \and Liang Li\textsuperscript{\rm 4} \and Weiming Hu\textsuperscript{\rm 1,3,5} \\
}
\institute{
    \textsuperscript{\rm 1}NLPR, Institute of Automation, Chinese Academy of Sciences, \textsuperscript{\rm 2}Megvii Technology\\
    \textsuperscript{\rm 3}School of Artificial Intelligence, University of Chinese Academy of Sciences\\
    \textsuperscript{\rm 4}Brain Science Center, Beijing Institute of Basic Medical Sciences\\
    \textsuperscript{\rm 5}CAS Center for Excellence in Brain Science and Intelligence Technology\\
    \texttt{wangshaoru2018@ia.ac.cn}; \texttt{\{jin.gao,wmhu\}@nlpr.ia.ac.cn},\\
    \texttt{\{lizeming\}@megvii.com}; \texttt{\{liang.li\}@aliyun.com} \\
}

\maketitle
\begin{abstract}
Self-supervised learning (SSL) has achieved promising downstream performance. However, when facing various resource budgets in real-world applications, it costs a huge computation burden to pretrain multiple networks of various sizes one by one. In this paper, we propose \textbf{D}iscriminative-SSL-based \textbf{S}limmable \textbf{P}retrained \textbf{Net}works (\textbf{DSPNet}), which can be trained at once and then slimmed to multiple sub-networks of various sizes, each of which faithfully learns good representation and can serve as good initialization for downstream tasks with various resource budgets. Specifically, we extend the idea of slimmable networks to a discriminative SSL paradigm, by integrating SSL and knowledge distillation gracefully. We show comparable or improved performance of DSPNet on ImageNet to the networks individually pretrained one by one under the linear evaluation and semi-supervised evaluation protocols, while reducing large training cost. The pretrained models also generalize well on downstream detection and segmentation tasks. Code will be made public.

\end{abstract}

\section{Introduction}

Recently, self-supervised learning (SSL) draws much attention to researchers where good representations are learned without dependencies on manual annotations. Such representations are considered to suffer less human bias and enjoy better transferability to downstream tasks. Generally, SSL solves a well-designed pretext task, such as image colorization \cite{image_colorization}, jigsaw puzzle solving \cite{jigsaw}, instance discrimination \cite{instance_discrimination}, and masked image modeling \cite{beit}. According to the different types of pretext tasks, SSL methods can be categorized into generative approaches and discriminative approaches. Discriminative approaches received more interest during the past few years, especially the ones following the instance discrimination pretext tasks \cite{mocov1, simclr, BYOL, mocov2, swav, simsiam}. Specifically, they learn similar and dissimilar representations from predefined positive pairs and negative pairs. BYOL \cite{BYOL} further demonstrates that such methods can also work well even without negative pairs. These pretraining approaches have shown superiority over their ImageNet-supervised counterpart in multiple downstream tasks. However, the pretraining is time-consuming and costs much computing resources, \eg, the pretraining of BYOL costs over 4000 TPU hours on the Cloud TPU v3 cores. 

In real-world applications, the resource budgets vary in a wide range for the practical deployment of deep networks. A single trained network cannot achieve optimal accuracy-efficiency trade-offs across different devices. It means that we usually need multiple networks of different sizes. A naive solution to apply pretraining and also meet such conditions is to pretrain them one by one and then fine-tune them to accomplish specific downstream tasks. However, their pretraining cost grows approximately linearly as the number of desired networks increases. It costs so many computing resources that this naive solution is far from being a practical way to make use of the favorable self-supervised pretraining. 

In this paper, a feasible approach is proposed to address this problem - developing slimmable pretrained networks that can be trained at once and then slimmed to multiple sub-networks of different sizes, each of which learns good representation and can serve as good initialization for downstream tasks with various resource budgets. To this end, We take inspiration from the idea of slimmable networks \cite{slimmable}, which can be trained at once and executed at different scale, and permit the network FLOPs to be dynamically configurable at runtime. Nonetheless, slimmable networks \cite{slimmable} and the subsequent works \cite{slimmablev2, bignas, cai2019once, mutualnet} all focus on specific tasks with supervision from manual annotations, while we concentrate on representation learning independent of manual annotations. To bridge this gap, we build our approach upon BYOL \cite{BYOL}, a representative discriminative SSL method. Specifically, as multiple networks of different sizes with good representations need to be pretrained at once, we thus construct them within a network family that are built with the same basic building blocks but with different widths and depths. We use \emph{desired networks} (DNs) to denote them. Different from the original design in BYOL that the online and target networks share the same architecture at each training iteration, our online network is specifically deployed as randomly sampled sub-networks, which include all the above constructed networks during pretraining. We also perform BYOL's similarity loss between the target branch and the sampled sub-networks in the online branch, and the target network is also updated by exponential moving averages of the online network. After training, DNs can be slimmed from the online network and each of them learns good representation. In this way, DNs are pretrained together by sharing weights, which can reduce large training cost compared with pretraining them one by one. We name the proposed approach as \textbf{D}iscriminative-SSL-based \textbf{S}limmable \textbf{P}retrained \textbf{N}etworks (\textbf{DSPNet}). Our contributions are summarized as follows:

\begin{itemize}
    \item We propose DSPNet, which can be trained at once and then slimmed to multiple sub-networks of various sizes, each of which learns good representation and serve as good initialization for downstream tasks with various resource budgets.
    \item We show that the slimmed networks achieve comparable or improved performance to individually pretrained networks under various evaluation protocols, \eg, the linear and semi-supervised evaluation on ImageNet~\cite{ImageNet} and the transferability evaluation to downstream detection and segmentation tasks on COCO~\cite{mscoco}, while large training cost is reduced.
    \item With extensive experiments, we show that DSPNet performs on par or better than previous distillation-based SSL methods, which can only obtain a single network with good representation by once training.
\end{itemize}

\section{Related Work}
\subsubsection{Self-supervised Learning.}
Recent self-supervised learning (SSL) approaches have shown prominent results by pretraining the models on ImageNet \cite{ImageNet} and transferring them to downstream tasks. It largely reduces the performance gap or even surpasses with respect to the supervised models, especially when adopting large encoders. Generally, SSL solves a well-designed pretext task. According to the types of pretext tasks, SSL methods can be categorized into generative approaches and discriminative approaches. Generative approaches focus on reconstructing original data based on the partial \cite{inpainting, beit, mae}, corrupted \cite{image_colorization, rotation_pred} or shuffled \cite{jigsaw} input, \etc. Discriminative approaches have drawn much attention in recent years, especially those based on the instance discrimination pretext tasks \cite{instance_discrimination}, which consider each image in a dataset as its own class. Among them, contrastive learning \cite{CL} methods achieve more promising performance, where the representations of different views of the same image are brought closer (positive pairs) while the representations of views from different images (negative pairs) are pushed apart. BYOL \cite{BYOL} further gets rid of negative pairs while preserving high performance, thanks to the additionally introduced predictor on top of the online network, namely non-contrastive method. Most of the above methods rely on large batch size and long training schedule \cite{BYOL,swav,mocov3}, which cost large computing resources.

\subsubsection{Knowledge Distillation.}
Knowledge distillation \cite{hinton2015distilling} generally aims to transfer knowledge from a well-trained model (teacher) to a compact network (student), which is also well-investigated in model compression \cite{compression}. The idea was first proposed by Hinton \etal \cite{hinton2015distilling}, where the mimic error of the student network towards a pretrained teacher network on output soft logits is adopted to guide the transfer of the knowledge. \mbox{FitNet}~\cite{fitnets} proposes to distill semantic information from intermediate features, and a mean-square-error-based hint loss is adopted. Besides, the channel attention \cite{zagoruyko2016at}, and relation of examples \cite{RelationKD} can also be considered as learned knowledge to be transferred. Moreover, knowledge distillation is widely adopted in various tasks besides classification tasks, \eg, object detection \cite{KD_Det} and semantic segmentation \cite{KD_Seg}. 
Some works also focus on utilizing self-supervised learning to boost the performance of task-specific distillation (\eg, image classification) under a fully supervised setting. For example, CRD \cite{CRD} combines self-supervision with knowledge distillation by a newly-introduced contrastive loss between teacher and student networks. In SSKD \cite{SSKD}, self-supervised learning is treated as an auxiliary task to help extract richer knowledge from a teacher network. 

\subsubsection{Self-supervised Learning with Distillation.}
Some research focuses on applying the aforementioned self-supervised methods to train lightweight networks, while the experimental results in \cite{SEED} show that the naive SSL methods do not work well for lightweight networks. To address this problem, knowledge distillation \cite{hinton2015distilling} is adopted by introducing a larger self-supervised trained teacher to help the lightweight networks learn good representations. For instance, CompRess \cite{CompRess} and SEED \cite{SEED} employ knowledge distillation to improve the self-supervised visual representation learning capability of small models, relying on MoCo \cite{mocov2} framework. A more flexible framework is proposed in \cite{DisCo} with contrastive learning adopted to transfer teachers' knowledge to smaller students. However, these two-stage methods all rely on the pretrained teachers and keep them frozen during distilling, which may be tedious and inefficient. Recently, some methods adopt online distillation \cite{digo, oss}, which recommends training the teacher and student at the same time. Our approach bears some similarity to them in that we also conduct SSL and knowledge distillation simultaneously. However, our method is aimed to obtain multiple networks of different sizes by training once to meet the demand of the practical deployment of deep networks under various resource budgets. Moreover, it can reduce large training cost.

\subsubsection{Slimmable Networks.}
Slimmable networks are a class of networks executable at different scale, which permit the network FLOPs to be dynamically configurable at runtime and enable the customers to trade off between accuracy and latency while deploying deep networks. The original version \cite{slimmable} achieves networks slimmable at different widths, and US-Nets \cite{slimmablev2} further extends slimmable networks to be executed at arbitrary widths, and also proposes two improved training techniques, namely the \emph{sandwich rule} and \emph{inplace distillation}. The subsequent works go beyond only changing the network width, \eg, additionally changing the network depth \cite{cai2019once}, kernel size \cite{bignas} and input resolutions \cite{mutualnet}. However, previous works all focus on specific tasks with supervision from manual annotations. Our approach extends them to SSL-based representation learning, to obtain slimmable pretrained models by once training. 

\section{Preliminaries}\label{sec:relation}
In this section, the mainstream discriminative SSL methods and slimmable networks are firstly summarized as preliminaries for our method. 

\subsection{Discriminative SSL}

Discriminative SSL methods generally adopt instance discrimination as pretext tasks. Among them, contrastive learning~\cite{CL, mocov1, mocov2} has attracted much attention.
Commonly, a paralleled architecture is adopted, in which augmented images are processed by two network branches, and contrastive loss is introduced \wrt predefined positive and negative pairs. One of the two branches usually adopts the \emph{stop-gradient} operation, \ie, not being updated directly based on backpropagation, while the other is optimized by the above loss\footnote{Some methods~\cite{barlowtwins} do not require the \emph{stop-gradient} operation, but rely on additional regularization, which is beyond the scope of this paper.}.
\begin{figure}[t]
    \centering
    \includegraphics[width=0.85\textwidth]{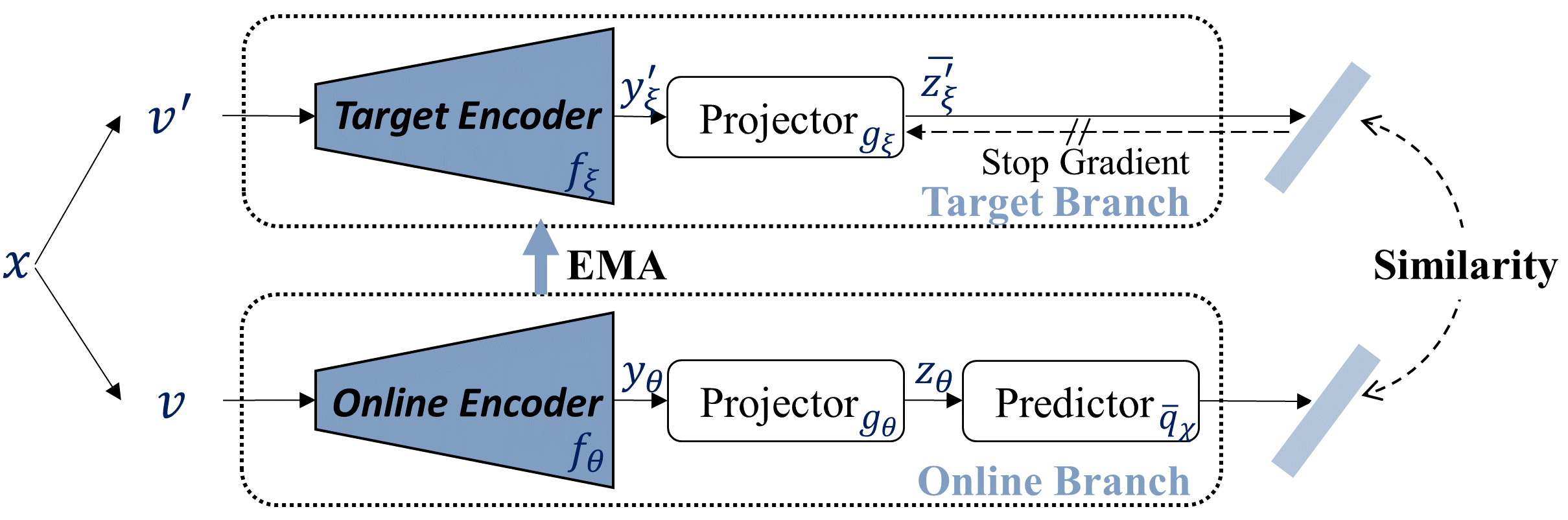}
    \caption{Architecture of BYOL~\cite{BYOL}. }
    \label{fig:byol}
\vspace{-10pt}
\end{figure}

This kind of SSL methods usually adopt different variants of the above architecture. For instance, MoCo \cite{mocov1, mocov2} consists of two paralleled encoders, \ie, a query encoder and a key encoder. The key encoder is utilized to construct a key dictionary as a queue, which is implemented as a momentum-based moving average of the query encoder, and InfoNCE \cite{cpc} loss is adopted. 
BYOL \cite{BYOL} further gets rid of the negative pairs, by introducing an additional predictor on top of the online network to prevent collapse, namely non-contrastive methods. Here we give a comprehensive description of BYOL, upon which our method is built. The architecture of BYOL is shown in \cref{fig:byol}. It consists of online and target networks. Given an image $x$ from datasets, two augmented views $v$ and $v'$ are obtained by applying different image augmentations respectively. From the first view $v$, the online network parameterized by $\theta$ first outputs a representation $y_\theta\triangleq f_\theta(v)$ and then a projection $z_\theta\triangleq g_\theta(y)$, and the target network parameterized by $\xi$ outputs a representation $y_\xi'\triangleq f_\xi(v')$ and the projection $z_\xi'\triangleq g_\xi(y')$ from the second view $v'$. A predictor $q_\chi$ parametered by $\chi$ is applied to the online branch, and mean squared error is adopted as prediction loss:
\begin{equation}
    L_{\theta,\xi,\chi}=\| \bar{q}_{\chi}(z_\theta)-\bar{z}_\xi'\|_2^2~,
\end{equation}
where $\bar{q}_{\chi}(z_\theta) \triangleq q_\chi(z_\theta) / \|q_\chi(z_\theta)\|_2$ and $\bar{z}_\xi' \triangleq z_\xi' / \|z_\xi'\|_2$ are the normalized version. Usually, a symmetric loss $\widetilde{L}_{\theta, \xi, \chi}$ is also computed by separately feeding $v'$ to the online network and $v$ to the target network. The parameters of the online branch $\theta$ and $\chi$ are finally updated by backpropagation from the total loss $L_{\mathrm{BYOL}}=L_{\theta, \xi, \chi}+\widetilde{L}_{\theta, \xi, \chi}$, and the parameters of the target branch $\xi$ are updated as an exponential moving average (EMA) of $\theta$. 

\subsection{Slimmable Networks}\label{sec:slimmable}

\begin{figure}[t]
    \centering
    \includegraphics[width=0.95\textwidth]{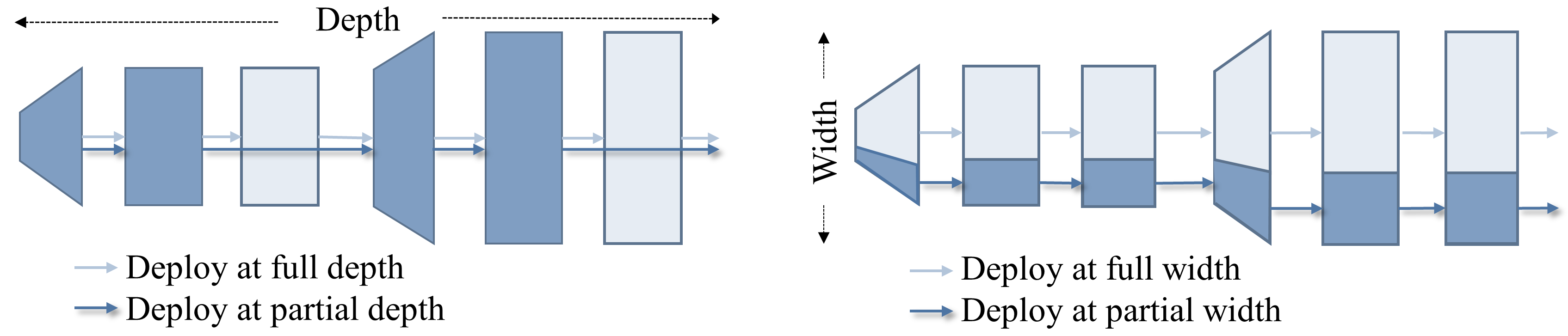}
    \caption{By dropping specific blocks (left) and inactivating parts of channels (right), one network can be slimmed to a smaller sub-network. We show them separately for clarity, while they could be applied together.}
    \label{fig:stochastic}
\end{figure}

Modern deep neural networks are mostly constructed by stacking a sequence of basic building blocks, \eg, \emph{BasicBlock} for ResNet18/34~\cite{resnet}, \emph{BottleNeck} for ResNet50/101/152~\cite{resnet}, \emph{InvertedResidual} for MobileNet~\cite{mobilenetv2,mobilenetv3}, \etc. By scaling the network width \cite{WRN, mobilenetv2, mobilenetv3}, depth \cite{resnet}, or both jointly \cite{efficientnet}, a family of networks can be obtained, which is known as model scaling. Specifically, augmenting the channel numbers of convolution operations or inserting blocks to different stages can lead to an expanded network for a lightweight network. Meanwhile, the lightweight network can be seen as a sub-network of the expanded network obtained by scaling, which makes it possible to share parameters between them. Then, the lightweight network can also be obtained by dropping specific blocks and inactivating parts of channels from the expanded network, as shown in \cref{fig:stochastic}.

Slimmable networks \cite{slimmable} are a class of networks based on the above model scaling strategy, which are executable at different scale. 
Several expanded networks from a lightweight network can be obtained with various scaling coefficients. All these networks (including the original lightweight network) are the sub-networks of the largest one among them. If we share the parameters of these expanded networks, we only need to attend to the parameters of the largest one and obtain the others by inactivating parts of channels or dropping a subset of blocks.

To train slimmable networks, a natural solution is to calculate the loss accumulated or averaged over all sub-networks. However, it is not practical or cost-effective if the number of sub-networks is large. To this end, \emph{sandwich rule} \cite{slimmablev2} is proposed, in which only the smallest, largest, and a few randomly sampled sub-networks are used in each iteration. After optimization, the slimmable networks can be deployed as any one of the candidate networks involved during the training phase. Besides, \emph{inplace distillation} \cite{slimmablev2} is also a useful training technique, in which the knowledge inside the largest network is transferred to sub-networks inplace in each iteration by simply adding distillation loss between them.


\section{Method}\label{sec:method}

Our goal is to obtain multiple networks of various sizes within a network family by training once, all of which have good representations after optimizing. We achieve it by integrating BYOL \cite{BYOL} and slimmable networks \cite{slimmable}. In this section, we introduce the proposed architecture and training procedure of our \textbf{D}iscriminative-SSL-based \textbf{S}limmable \textbf{P}retrained \textbf{N}etworks (\textbf{DSPNet}).

\subsection{DSPNet}
\begin{figure}[t]
    \centering
    \includegraphics[width=0.85\textwidth]{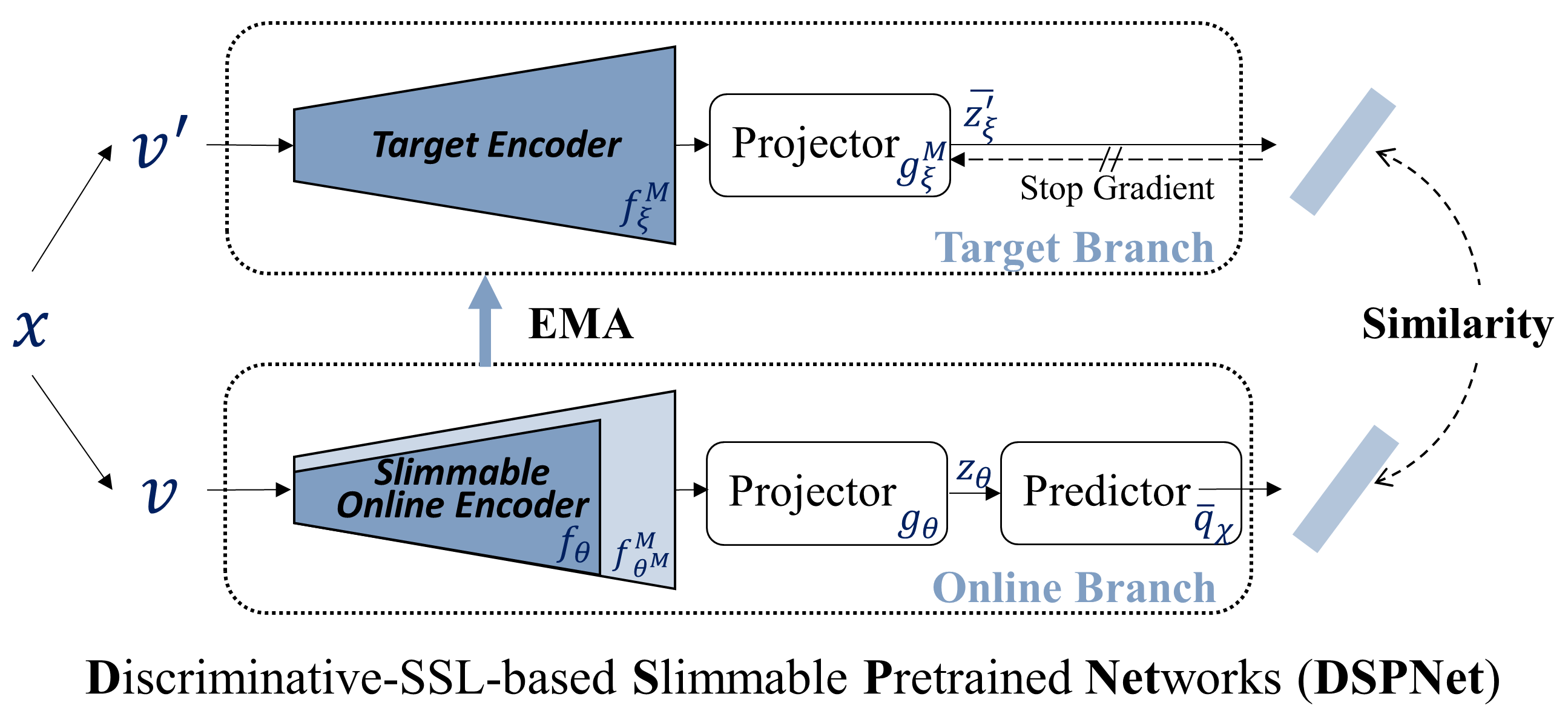}
    \caption{Architecture of our DSPNet. Based on BYOL \cite{BYOL}, our online encoder is deployed as randomly sampled sub-networks at each training iteration, while the target encoder is always deployed at full depth and width. After training, the optimized online encoder can be slimmed to multiple networks of different sizes, and all of them have good representations.}
    \label{fig:arch}
\end{figure}

The overall architecture of our approach, namely DSPNet, is shown in ~\cref{fig:arch}. It largely follows BYOL~\cite{BYOL}, except for the encoder of the online branch. Unlike the original design in BYOL, in which the target encoder itself is the desired network with good representations after training, our desired networks are several slices of the online encoder\footnote{Since the target network is obtained by exponential moving average of the online network, it should converge to the same status as the online encoder. In our experiments, we find that the representations of the sliced networks from the target and online branches can achieve comparable results.}. They are sub-networks by removing specific channels and blocks from the whole online network, referred to as \emph{desired networks} (DNs) in this paper. 
We also adopt the \emph{sandwich rule} \cite{slimmable}. Specifically, at each training iteration, we sample $n$ sub-networks, involving the smallest, largest and randomly sampled ($n$-2) sub-networks, then calculate the loss for each of them, and finally apply gradients back-propagated from the accumulated loss of them. The projector and predictor along with the loss function all follow the original design in BYOL~\cite{BYOL}, except for the first linear layer in the projector of the online branch, because it should have an alterable input channel to adapt to the output representations with different dimensions from the online encoder. As for the target branch, the encoder has the same architecture as the whole online network and is always deployed at full width (\ie, all channels are activated) and depth (\ie, all blocks are maintained).

More specifically, for one sampled online network, which is defined by a set of weights $\theta$, we denote its encoder as $f_\theta \sim \mathcal{F}$, and then the projection of the representation from the online encoder is calculated as 
\begin{equation}
    z_\theta\triangleq g_\theta(f_\theta(v))
\end{equation}
for the first view $v$, where $g_\theta$ is the projector. It is noteworthy that if the whole online encoder is defined as $f^M$ parameterized by $\theta^M$, then the parameters of any sub-network are a subset of $\theta^M$. For the second view $v'$, the projection of the representation from the encoder of the target branch is calculated as 
\begin{equation}
    z_\xi'\triangleq g_\xi^M(f_\xi^M(v')),
\end{equation}
where $f_\xi^M$ and $g_\xi^M$ are the encoder and projector of the target branch parameterized by $\xi$, which are deployed at full size the same as the whole online network, \ie, $|\xi|=|\theta^M|$. Then, our prediction loss is defined as
\begin{equation}
    L_{\theta, \xi, \chi}=\| \bar{q}_{\chi}(z_\theta)-\bar{z}_\xi'\|_2^2~,
\end{equation}
where $q_\chi$ is the predictor parameterized by $\chi$ in the online branch, and $\bar{q}_{\chi}(z_\theta)$ and $\bar{z}_\xi'$ are the normalized version of $q_\chi(z_\theta)$ and $z_\xi'$. 

We compute the loss by taking an unweighted sum of all training losses of sampled networks:
\begin{equation}\label{equ:loss}
    L_{\Theta, \xi, \chi}=\sum_{i=1}^{n}\| \bar{q}_{\chi}(z_{\theta_i})-\bar{z}_\xi'\|_2^2~,
\end{equation}
where $\Theta=\theta_1\bigcup \theta_2 \bigcup \dots \bigcup \theta_n$ and $\theta_i$ is the parameters of the $i$-th sampled online network. Since the largest sub-network (the whole one) is always sampled in the online branch following the \emph{sandwich rule}, we can find that $\Theta=\theta^M$. We also symmetrize the loss in \cref{equ:loss} by interchanging $v$ and $v'$ to compute $\widetilde{L}_{\theta^M, \xi, \chi}$ following the practice in BYOL. The total loss is $L_{\theta^M, \xi, \chi}^{\mathrm{DSPNet}}=L_{\theta^M, \xi, \chi}+\widetilde{L}_{\theta^M, \xi, \chi}$. $\theta^M$ along with $\chi$ are updated based on the gradients from backpropagation, and $\xi$ is updated by a momentum-based moving average of $\theta^M$ with hyper-parameter $\tau$, \ie, \begin{align}
    &\theta^M,\chi \leftarrow \mathrm{optimizer}(\theta^M,\chi, \nabla_{\theta^M\!\!,\chi}~L_{\theta^M, \xi, \chi}^{\mathrm{DSPNet}})~,\\
    &\xi \leftarrow \tau \xi + (1-\tau)\theta^M~.
\end{align}

%

Since we can accumulate back-propagated gradients of all sampled networks and do not need to always keep all intermediate results, the training procedure is memory-efficient, with no more GPU memory cost than pretraining individual networks. Moreover, at each training iteration, $\bar{z}_\xi'$ is shared for all sampled online encoders, which means it is also time-efficient.
After pretraining, the \emph{desired networks} (DNs) can be slimmed from the whole online encoder, all of which are expected to have good representations.


\subsection{Behaviors Behind the Training of DSPNet}
Overall, our online encoder varies within the predefined DNs during pretraining. 
When the whole network is sampled, our training behaves like BYOL \cite{BYOL}. It is essential for our approach to achieve good representations. While for the other sampled sub-networks, the supervision is built upon the similarity between the representations from the target branch (with the same size as the whole online network) and this partially activated online network, which can be viewed as knowledge distillation so that the representation capability is transferred and compressed to the sub-networks. It is noteworthy that, different from the \emph{inplace distillation} of slimmable networks \cite{slimmablev2}, our teacher is the target network rather than the whole online network, which is a momentum-based moving average of the whole online network and severed as a stable and reliable teacher. 

In this view, our training is jointly optimized towards self-supervised learning and knowledge distillation, and thus contributes to simultaneous representation learning and model compression. 

\section{Experiments}

We train our DSPNet on ImagetNet ILSVRC-2012 \cite{ImageNet} without labels. Then we evaluate the representations of our obtained DNs on the validation set of ImageNet, including linear evaluation protocol and semi-supervised evaluation protocol. The transferability to the downstream tasks, such as object detection and instance segmentation on COCO \cite{mscoco}, is also investigated to study the generalization.

\subsection{Comparisons with Individually Pretrained Networks}
One of the advantages of our pretraining method is that we can obtain multiple \emph{desired networks} (DNs) by training once, which is cost-efficient compared with individually pretraining them one by one. Thus, experimental comparisons with individually pretrained networks are carried out in this section.
\subsubsection{Implementation Details.} \label{sec:details}
We mainly follow the settings in BYOL~\cite{BYOL}. The same set of image augmentations as in BYOL \cite{BYOL} is adopted. Our DNs are MobileNet v3 \cite{mobilenetv3} with different widths, \ie, $[0.5, 0.625, 0.75, 0.875, 1.0]\times$. The projector and predictor are the same as in BYOL \cite{BYOL}. We also adopt momentum BN \cite{m2t} for further accelerating. We use the LARS \cite{lars} optimizer with a cosine decay learning rate schedule \cite{sgdr} and 10-epoch warm-up period. We set the base learning rate to 0.2, scaled linearly \cite{imagenet_in_1_hour} with the batch size (\ie, 0.2$\times$BatchSize/256). The weight decay is $1.5\cdot10^{-6}$.
The EMA parameter $\tau$ starts from $\tau_{base} = 0.996$ and is gradually increased to $1.0$ during training. We train for 300 epochs with a mini-batch size of 4096. The number of sampled sub-networks per iteration $n$ is 4.
Besides, we constrain the online encoder to be deployed at full scale at the first 10 warm-up epochs, for robust training. As our baselines, we individually pretrain these models (MobileNet v3 at widths $[0.5, 0.625, 0.75, 0.875, 1.0]\times$) with BYOL \cite{BYOL}. The same training recipe as ours is adopted.

\subsubsection{Linear Evaluation on ImageNet.}\label{sec:sdp-lp}
\begin{figure*}[t]
    \centering
    \begin{minipage}[t][][b]{0.48\textwidth}
        \centering
        \includegraphics[width=1.0\textwidth]{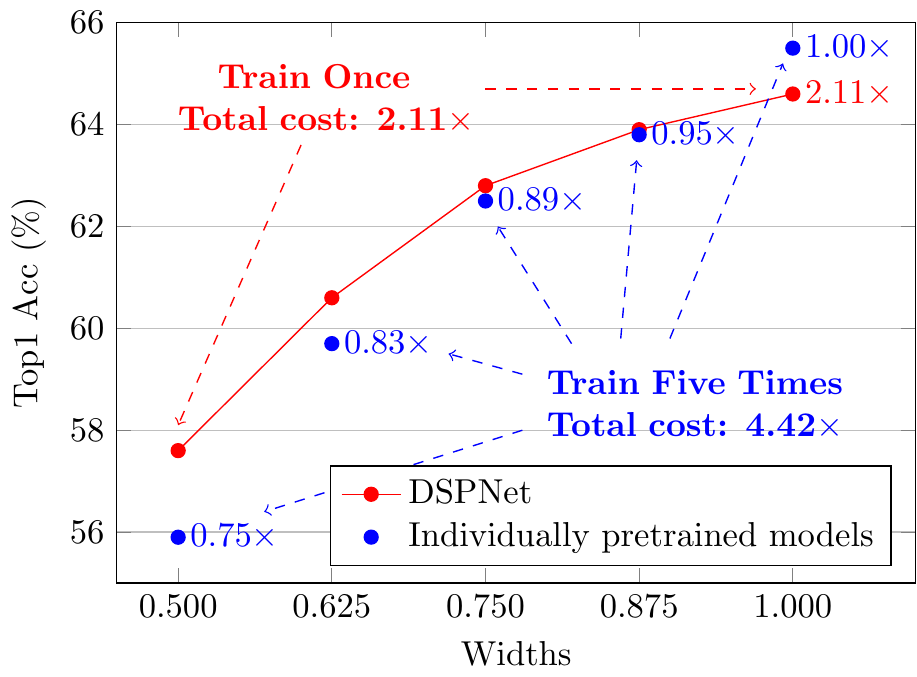}
        \vspace{-10pt}
        \caption{\textbf{Top1 accuracy (\%) on ImageNet under linear evaluation} for DNs slimmed from our DSPNet and individually pretrained models. Relative training time is also reported.}
        \label{fig:sdp-lp}
    \end{minipage}\hspace{2mm}
    \begin{minipage}[t][][b]{0.48\textwidth}
        \centering
        \includegraphics[width=1.0\textwidth]{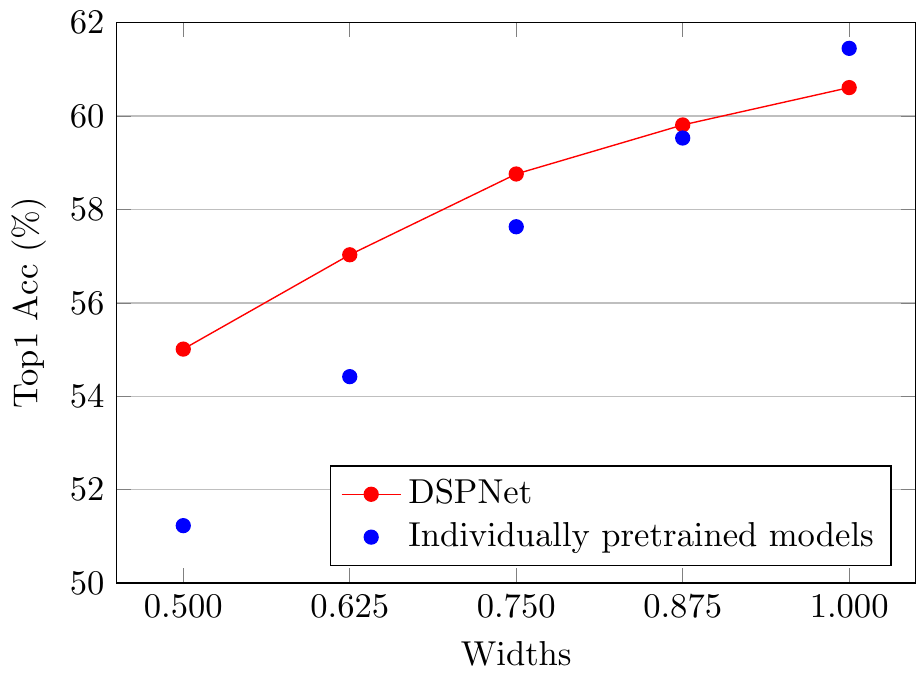}
        \vspace{-10pt}
        \caption{\textbf{Semi-supervised learning on ImageNet} using 10\% training examples for DNs slimmed from our DSPNet and individually pretrained models. Top1 accuracy (\%) is reported.}
        \label{fig:sdp-semi}
    \end{minipage}\hspace{2mm}
\vspace{-1pt}
\end{figure*}
Following the ImageNet~\cite{ImageNet} linear evaluation protocol \cite{BYOL} as detailed in \cref{sec:a-linear}, we train linear classifiers on top of the frozen representations with the full training set, and then report the top1 accuracy on the validation set. Note that the classifier is trained individually for each DN. As shown in \cref{fig:sdp-lp}, our method achieves comparable performance to the individually pretrained models. Specifically, for smaller DNs, our method achieves minor improvement, while showing slightly inferior results only for the largest one. 

Here, we provide one possible explanation for the above observations by investigating the working mechanism of our DSPNet. We first consider the individually pretrained baseline with MobileNet v3 at width $1.0\times$ as the encoder for both the target and online branches, which achieves 65.5\% top1 accuracy under linear evaluation. The introducing of optimizing sampled sub-networks for the online encoder during the pretraining of our DSPNet can be viewed as an additional regularization on the full-size online encoder. This urges the good representation to be compressed to the DNs. In other words, such a training scheme makes a compromise between slimmablity and good representation for the full-size online encoder. The former accounts for a good imitation for the DNs to the whole one, and the latter for the qualified target network as a teacher, both of which contribute to powerful DNs. Hence, smaller DNs benefit from the knowledge distillation from the teacher and achieve better performance than individually pretrained ones, while the largest one has to accommodate those DNs as its sub-networks since no explicit supervision is provided in SSL, thus showing slightly inferior results.

Furthermore, to evaluate the efficiency of our method, we report the relative pretraining time \wrt the individually pretrained MobileNet v3 at width $1\times$, which is also shown in \cref{fig:sdp-lp}. It is evaluated on 16 V100 32G GPUs. The pretraining of our DSPNet spends $2.11\times$ time \wrt the above baseline, while it totally costs $4.42\times$ time to pretrain all the DNs one by one. About a half of training cost can be reduced by our method. 


\begin{figure*}[t]
    \centering
    \begin{minipage}[t][][b]{0.48\textwidth}
        \centering
        \includegraphics[width=1.0\textwidth]{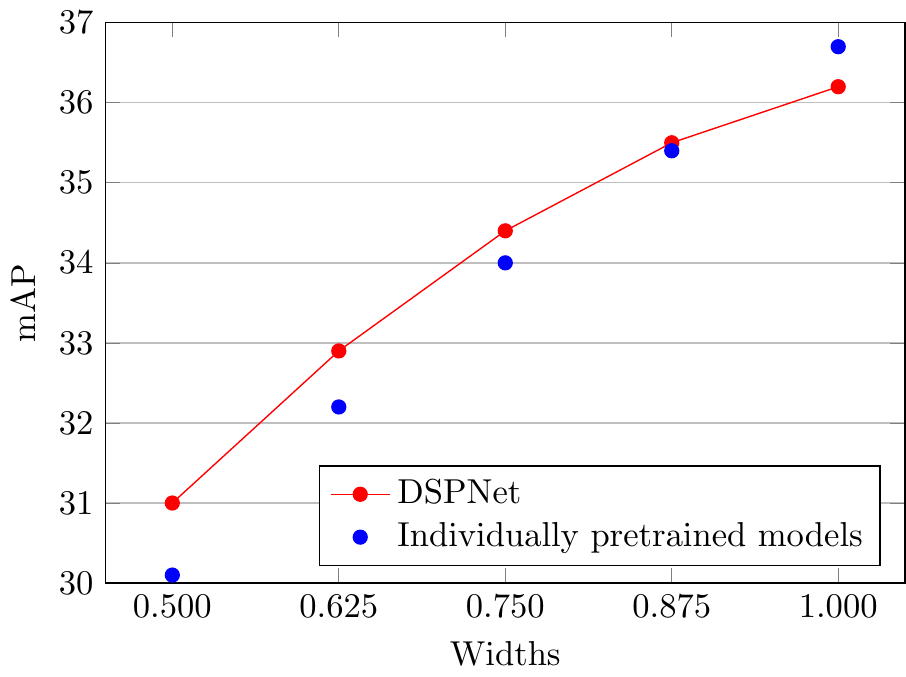}
    \end{minipage}\hspace{2mm}
    \begin{minipage}[t][][b]{0.48\textwidth}
        \centering
        \includegraphics[width=1.0\textwidth]{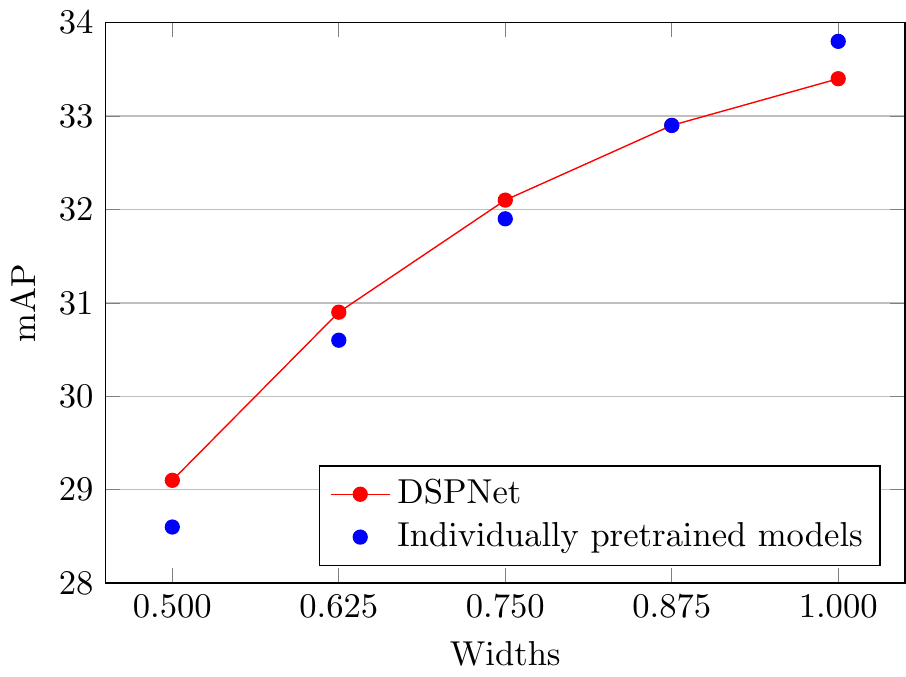}
    \end{minipage}\hspace{2mm}
\caption{\textbf{COCO object detection (left) and segmentation (right)} based on Mask R-CNN. mAP is reported for both DNs slimmed from our DSPNet and individually pretrained models.}
\label{fig:sdp-det}
\vspace{-1pt}
\end{figure*}


\subsubsection{Semi-supervised Evaluation on ImageNet.}\label{sec:sdp-semi}
We also evaluate the representations on a classification task with a small subset of ImageNet's training set available for fine-tuning them, \ie, semi-supervised evaluation, as shown in \cref{fig:sdp-semi}. We fine-tune the pretrained models by only utilizing 10\% of the labeled training data, including both multiple DNs slimmed from our DSPNet and individually pretrained models. Similar phenomena are observed as in linear evaluation. Higher accuracy than individually pretrained models is achieved except for the largest DN. More details can be found in \cref{sec:a-semi}.

\subsubsection{Transferability to Other Downstream Tasks.}\label{sec:sdp-det}

We use Mask R-CNN~\cite{mask_rcnn} with FPN \cite{fpn} to examine whether our representations generalize well beyond classification tasks (see \cref{sec:a-transfer} for more details). We fine-tune on COCO \textit{train2017} set and evaluate on \textit{val2017} set. All the model parameters are learnable. We follow the typical 2$\times$ training strategy \cite{detectron2}. As shown in \cref{fig:sdp-det}, our bounding box AP and mask AP both outperform the individually pretrained counterparts for most models.

\subsubsection{Applying to Other Networks.}\label{sec:sdp-resnet}

\begin{wrapfigure}{r}{0.5\textwidth}
    \centering
    \vspace{-20pt}
    \includegraphics[width=0.5\textwidth]{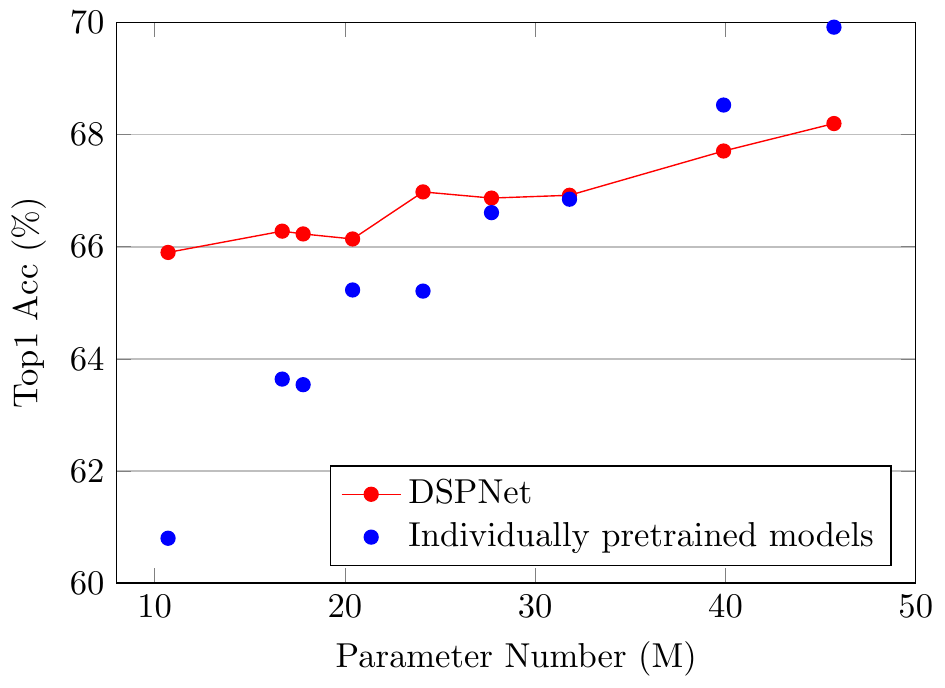}
    \caption{\textbf{Applying DSPNet to ResNet.}}
    \label{fig:multiple}
\vspace{-5pt}
\end{wrapfigure}

Our method is generally applicable to most representative network architectures. We further evaluate our method on ResNet \cite{resnet}. We construct several DNs based on the \emph{BasicBlock} building blocks. Specifically, the channel numbers for the $4$ stages are sampled from $\{[64,128,256,512]$, $[80,160,320,640]$, $[96,192,384,768]\}$ and the block numbers for the $4$ stages are sampled from $\{[2,2,2,2]$, $[2,3,4,3]$, $[3,4,6,3]\}$, resulting in $9$ networks as DNs. The smallest one among them is the well-known ResNet18, and the largest one is ResNet34 at width $1.5\times$. The top1 accuracy of linear evaluation \wrt their parameter numbers is shown in \cref{fig:multiple}. The performance of these networks individually pretrained with BYOL \cite{BYOL} is also evaluated. The same training configurations as those of MobileNet v3 are adopted, except that the images are resized to $128\times128$ for faster training. 

It can be drawn that for most smaller models, our DNs slimmed from the whole network outperforms those individually pretrained ones, while falling behind for a few larger ones. We attribute it to the conjecture that the whole online network makes a compromise between slimmablity and good representation.

\subsection{Ablation Study}
\begin{figure*}[t]
    \centering
    \begin{minipage}[t][][b]{0.48\textwidth}
        \centering
        \includegraphics[width=1.0\textwidth]{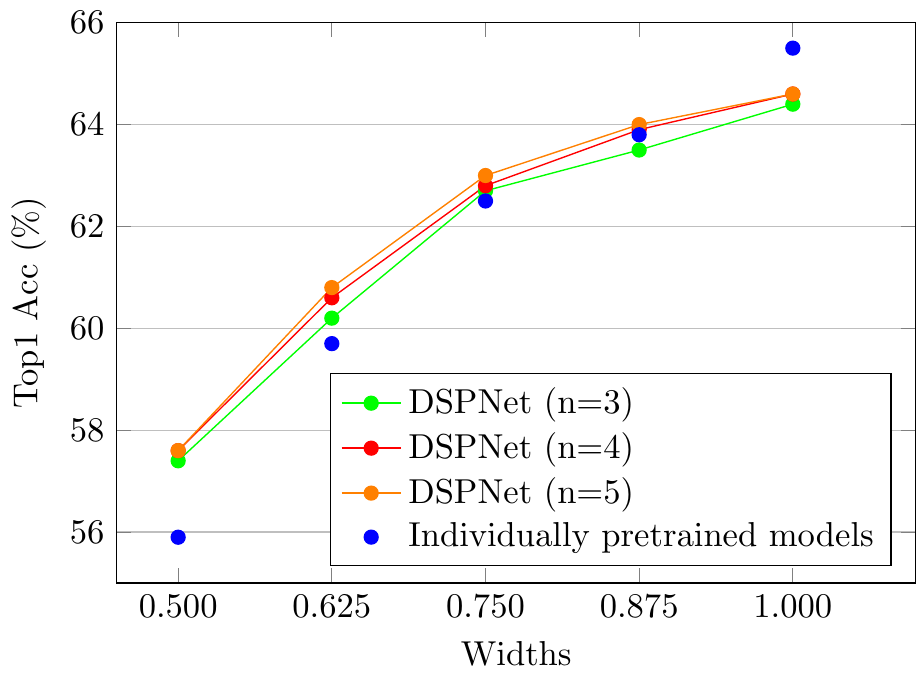}
    \end{minipage}\hspace{2mm}
    \begin{minipage}[t][][b]{0.48\textwidth}
        \centering
        \includegraphics[width=1.0\textwidth]{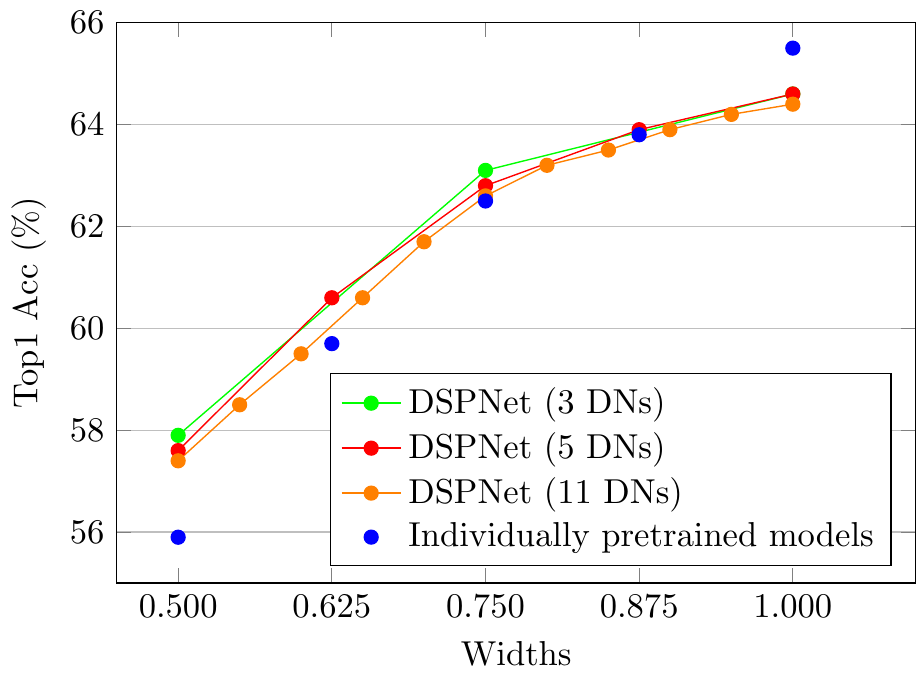}
    \end{minipage}\hspace{2mm}
\caption{\textbf{Ablation study} under linear evaluation. We examine the effect of the number of sampled sub-networks per iteration (left), and the number of pre-defined DNs (right). Top1 accuracy is reported for both.}
\label{fig:ablation}
\vspace{-1pt}
\end{figure*}
\begin{table*}[h!]
\centering
\begin{subtable}[c]{1.0\textwidth}
    \centering
    \setlength{\tabcolsep}{5pt}
    \renewcommand{\arraystretch}{0.9} 
    \begin{tabular}{cccccc}
        \multirow{2}{*}{\textbf{Methods}} &  \multicolumn{2}{c}{\textbf{Teacher}} &  \multicolumn{3}{c}{{\textbf{Student} (ResNet18)}} \\
        \cmidrule(lr){2-3} \cmidrule(lr){4-6}
        & Methods & Top1 & kNN & Top1 & Top5 \\
        \bottomrule
        \multicolumn{6}{c}{\emph{Supervised}} \\
        & - & - & \multicolumn{3}{c}{69.5}\\
        \bottomrule
        \multicolumn{6}{c}{\emph{Self-supervised}}\\
        MoCo-v2 \cite{mocov2} & - & - & 36.7 & 52.2 & 77.6 \\
        BYOL \cite{BYOL} & - & - & 50.2 & 60.8 & 83.6 \\
        \bottomrule
        \multicolumn{6}{c}{\emph{Self-supervised with Distillation}}\\
        SEED \cite{SEED} & R50 (MoCo-v2\cite{mocov2}+200e) & 67.4 & 43.4 & 57.9 & 82.0 \\
        & R50$\times$2 (SwAV\dag\cite{swav}+400e) & 77.3 & 55.3 & 63.0 & 84.9 \\
        CompRess \cite{CompRess} & R50 (MoCo-v2\cite{mocov2}+800e) & 70.8 & 53.5 & 62.6 \\
        & R50 (SwAV\cite{swav}+800e) & 75.6 & 56.0 & 65.6 & - \\
        DisCo \cite{DisCo} & R50 (MoCo-v2\cite{mocov2}+200e) & 67.4 & - & 60.6 & 83.7\\ 
        & R50$\times$2 (SwAV\dag\cite{swav}+400e) & 77.3 & - & 65.2 & \textbf{86.8}\\
        Ours & expanded & - & \textbf{56.1} & \textbf{65.9} & 86.4\\
    \toprule
    \end{tabular}
    \caption{
    \textbf{kNN and linear evaluation comparisons} on ImageNet. R50 is short for ResNet50 \cite{resnet}, and R50$\times$2 represents a R50 at width $2\times$. $\dag$ indicates using multi-crop strategy \cite{swav}. 
    }
    \label{tbl:resnet-lp}
\end{subtable}
\\
\begin{subtable}[h]{0.49\textwidth}
    \centering
    \setlength{\tabcolsep}{3.3pt}
    \renewcommand{\arraystretch}{1.06} 
    \begin{tabular}{ccccccccccc}
    \multirow{2}{*}{\makecell[c]{\textbf{Methods}\\(ResNet18)}} & \multicolumn{2}{c}{\textbf{1\%}} & \multicolumn{2}{c}{\textbf{10\%}} \\
    \cmidrule(lr){2-3} \cmidrule(lr){4-5} \\[-2.5ex]
    & Top1 & Top5 & Top1 & Top5 \\
     \midrule
    MoCo-v2~\cite{mocov2}& 31.1 & 54.5 & 47.2 & 73.1 \\
    BYOL~\cite{BYOL} & 42.0 & 67.4 & 56.1 & 81.4 \\
    \midrule
    SEED \cite{SEED} & 43.5 & - & 54.9 & - \\
    DisCo \cite{DisCo} & 47.5 & - & 54.6 & - \\
    \midrule
    Ours & 47.1 & 71.8 & 59.7 & 82.0 \\
    Ours (fine-tuning) & \textbf{49.2} & \textbf{73.7} & \textbf{62.5} & \textbf{83.6}\\
    
    \toprule
    \end{tabular}
    \caption{
    \textbf{Semi-supervised evaluation} on ImageNet. Top1 and Top5 accuracy are reported.
    }
    \label{tbl:resnet-semi}
\end{subtable}\hfill
\begin{subtable}[h]{0.49\textwidth}
    \centering
    \setlength{\tabcolsep}{1.0pt}
    \renewcommand{\arraystretch}{1.1} 
    \resizebox{1.0\textwidth}{!}{
    \begin{tabular}{ccccccccccc}
    \multirow{2}{*}{\makecell[c]{\textbf{Methods}\\(ResNet18)}} & \multicolumn{3}{c}{\textbf{COCO Obj. Det.}} & \multicolumn{3}{c}{\textbf{COCO Inst. Segm.}} \\
    \cmidrule(lr){2-4} \cmidrule(lr){5-7} \\[-3.0ex]
    & AP$^{\text{bb}}$  & AP$^\text{bb}_{50}$  & AP$^\text{bb}_{75}$ & AP$^{\text{mk}}$  & AP$^\text{mk}_{50}$  & AP$^\text{mk}_{75}$ \\
     \\[-3.5ex]
     \midrule
     \emph{Random Init.} & 34.7 & 54.3 & 37.7 & 31.8 & 51.2 & 34.0
    \\
     \midrule 
    \emph{Supervised-IN.} & 37.1 & 57.3 & 40.0 & 34.0 & 54.3 & 36.4
    \\
    \midrule
    MoCo-v2~\cite{mocov2}& 36.6	& 56.7 & 39.7 & 33.4 & 53.8 & 35.8
     \\
    BYOL~\cite{BYOL} & 36.8	& 57.1 & 39.7 & 33.6 & 54.2 & 35.7 \\
    \midrule
    SEED \cite{SEED} & 37.1 & 57.6 & 40.0 & 34.0 & 54.5 & \textbf{36.3} \\
    CompRess \cite{CompRess} & 36.9 & 57.3 & 39.8 & 33.7 & 54.3 & 36.0
    \\
    \midrule
    Ours & \textbf{37.2} & \textbf{57.8} & \textbf{40.4} & \textbf{34.1} & \textbf{54.9} & \textbf{36.3}
    \\
    \toprule
    \end{tabular}
    }
    \caption{
    \textbf{COCO \cite{mscoco} object detection and segmentation} based on Mask R-CNN \cite{mask_rcnn}. 
    }
    \label{tbl:resnet-det}
\end{subtable}
\caption{\textbf{Comparisons with SSL methods with distillation.} We compare the smallest DN slimmed from our DSPNet with the models trained by SSL methods with distillation \cite{SEED,DisCo,CompRess}. They are all with ResNet18 \cite{resnet} network architecture. Our method achieves comparable or better results to these methods under various evaluation protocol. }
\end{table*}
We now explore the effects of the hyper-parameters of our method. For concise ablations, all experiments are conducted based on MobileNet v3 \cite{mobilenetv3} and the pretrained models are evaluated under the linear evaluation protocol.
\subsubsection{Number of Sampled Sub-networks Per Iteration $n$.}
We first study the number of sampled sub-networks per training iteration. It is worth noting that larger $n$ leads to more training time. We train three kinds of DSPNet models with $n$ equal to $3$, $4$, or $5$. As shown in the left of \cref{fig:ablation}, the models pretrained with $n=4$ achieve better results than $n=3$ under linear evaluation, but further increasing to $n=5$ does not lead to significant improvement. Thus, we adopt $n=4$ in all our experiments by default.

\subsubsection{Number of DNs.}
We then investigate the effect of the number of DNs. The more DNs, the more pretrained models can be obtained by once training. As shown in the right of \cref{fig:ablation}, we have pretrained 3-switch, 5-switch, and 11-switch DSPNet. They all show similar performance, demonstrating the scalability of our method.

\subsection{Comparisons with SSL Methods with Distillation}

We notice that there are some works \cite{SEED, CompRess, DisCo} focusing on the self-supervised training of lightweight networks with knowledge distillation. From some perspectives, our method can be viewed as an online distillation method by training the teacher (the target network) and students (sub-networks of the online network) simultaneously. Thus we compare our method with them by only picking the smallest DN slimmed from our DSPNet. The experiments are conducted based on ResNet. We directly use the models introduced in \cref{sec:sdp-resnet}, where the smallest DN is ResNet18. 

First, we perform comparisons under the k-nearest-neighbor (kNN) classification and linear evaluation on ImageNet \cite{ImageNet}. As shown in \cref{tbl:resnet-lp}, we achieve better results than other distillation-based SSL methods \cite{SEED,CompRess, DisCo}. It is worth noting that our method does not rely on well-pretrained teachers, and additional models with good representations can also be obtained at the same time. Thus, the training cost can be largely saved. 

Second, we evaluate the pretrained models under semi-supervised evaluation. Following previous works \cite{SEED}, we train a linear classifier on top of the frozen representation by only utilizing 1\% and 10\% of the labeled ImageNet's training data. As shown in \cref{tbl:resnet-semi}, the smallest DN slimmed from our DSPNet achieves 47.1\% and 59.7\% top1 accuracy respectively. Our method outperforms previous approaches on 10\% semi-supervised linear evaluation by a large margin. Higher accuracy is achieved if we further perform fine-tuning on this semi-supervised classification task, in which the parameters in backbone can be optimized along with the classifier. Details can be found in \cref{sec:a-semi}.

Finally, we transfer the pretrained models to downstream tasks, \ie, object detection and segmentation tasks on COCO \cite{mscoco}. We also use Mask R-CNN \cite{mask_rcnn} with FPN \cite{fpn} as in \cref{sec:sdp-det} by only replacing the backbone with ResNet18. As shown in \cref{tbl:resnet-det}, our bounding box AP and mask AP both outperform the naive SSL counterparts (\eg, MoCo-v2 \cite{mocov2} and BYOL \cite{BYOL}), and also the distillation-based methods (\eg, SEED \cite{SEED} and CompRess \cite{CompRess}), which rely heavily on large pretrained teachers. Our method also shows comparable results to the supervised baseline (\textit{Supervised-IN.} entry in \cref{tbl:resnet-det}) with ResNet18 \cite{resnet}, which relies on the labels of ImageNet dataset \cite{ImageNet}.

\section{Conclusions}
In this paper, we have proposed Discriminative-SSL-based Slimmable Pretrained Networks (DSPNet), which can be trained at once and then slimmed to multiple sub-networks of various sizes, and experimental results show that each of them learns good representation and can serve as good initialization for downstream tasks, \eg, semi-supervised classification, detection, and segmentation tasks. Our method perfectly meets the practical use where multiple pretrained networks of different sizes are demanded under various resource budgets, and large training cost can be reduced. 

\section{Acknowledgements}
This work was supported by the National Key R\&D Program of China (Grant No. 2018AAA0102802, 2018AAA0102800), the Natural Science Foundation of China (Grant No. 61972394, 62036011, 62192782, 61721004, U2033210, 62172413), the Key Research Program of Frontier Sciences, CAS (Grant No. QYZDJ-SSW-JSC040), the China Postdoctoral Science Foundation (Grant No. 2021M693402). Jin Gao was also supported in part by the Youth Innovation Promotion Association, CAS.

\clearpage
%
%
\small
\bibliographystyle{splncs}
\bibliography{egbib}
\clearpage
\appendix
\setcounter{table}{0}
\renewcommand{\thetable}{A\arabic{table}}
\setcounter{equation}{0}
\renewcommand{\theequation}{A\arabic{equation}}
\setcounter{figure}{0}
\renewcommand{\thefigure}{A\arabic{figure}}
\pagenumbering{Roman}



\title{Supplementary Materials: \\DSPNet: Towards Slimmable Pretrained Networks
based on Discriminative Self-supervised Learning}
\titlerunning{DSPNet}
\authorrunning{S. Wang et al.}

\author{}
\institute{}
\maketitle

This is a supplementary material for our paper ``DSPNet: Towards Slimmable Pretrained Networks based on Discriminative Self-supervised Learning". Due to spatial constraints in the main paper, some more detailed explanations are moved to this supporting document. In specific, we provide more \textbf{implementation details} for our method to facilitate the reproduction of our results, and additionally show that our obtained full-size online encoder based on DSPNet can also sever as an \textbf{ideal initialization for the supervised training of slimmable networks} \cite{slimmable, slimmablev2}, which is better than the random initialization or BYOL \cite{BYOL} pretraining.

\section{Implementation Details}

\subsection{Architecture}\label{sec:a-arch}
We use MobileNet v3 \cite{mobilenetv3} at width $1\times$ as the architecture of the whole online encoder. A \emph{multi-layer perceptron} (MLP) network, which consists of a linear layer with output size 4096 followed by batch normalization \cite{BN}, rectified linear units (ReLU) \cite{ReLU}, and a final linear layer with output dimension 256, is adopted as the projectors of both the online and target branches, \ie, $g_\theta$ and $g_\xi^M$ respectively. The predictor $q_\chi$ uses the same architecture as $g_\theta$. The above architecture settings are also applied to ResNet-based experiments without modifications. 


\subsection{Linear and kNN Evaluation}\label{sec:linear}\label{sec:a-linear}
We provide more details of our linear evaluation in \cref{fig:sdp-lp} and \cref{tbl:resnet-lp}.
We largely follow the setting in BYOL \cite{BYOL}, which includes training a linear classifier on top of the frozen representation, \ie, without updating the backbone feature extraction network parameters or the batch statistics. We add an additional BN \cite{BN} layer on top of the linear classifier. In our setting, it affects the performance little, but makes the training less sensitive to the learning rate. The learning rate for training the linear classifiers with respect to MobileNet v3 \cite{mobilenetv3} and ResNet \cite{resnet} is 0.2 with a batch size 256. SGD with a momentum of 0.9 is adopted. We train the classifiers for 80 epochs in total with a cosine decay learning rate schedule \cite{sgdr}. For kNN evaluation in \cref{tbl:resnet-lp}, a sample is classified by taking the most frequent label of its K-nearest neighbors (K = 20) based on the same frozen representation as the above linear evaluation. 

\subsection{Semi-supervised Evaluation}\label{sec:a-semi}
For the semi-supervised evaluation, the training set is altered to 1\% or 10\% subset. By default, we adopt fine-tuning evaluation (\cref{fig:sdp-semi}, and the entry marked with \emph{fine-tuning} in \cref{tbl:resnet-semi}), where the parameters in both the backbone and the classifier are all trainable. And distinct learning rates are applied to the backbone and the classifier following \cite{swav}, \ie, 0.001 and 0.02 respectively. The same training settings are adopted for MobileNet v3 \cite{mobilenetv3} and ResNet \cite{resnet}. 

Furthermore, to make fair comparisons to SSL methods with distillation \cite{SEED,DisCo} as shown in \cref{tbl:resnet-semi}, we adopt the same semi-supervised linear evaluation protocol as them by training a linear classifier on top of the frozen representation of ResNet18 slimmed from our DSPNet. Specifically, the configurations largely follow the linear evaluation in \cref{sec:a-linear}, except that the learning rate for the classifier is set to 0.02 and the number of training epochs is reduced to 50. The performance of semi-supervised linear evaluation for SEED \cite{SEED} and Disco \cite{DisCo} in \cref{tbl:resnet-semi} is directly borrowed from their papers, both of which employ a large pretrained R152 \cite{resnet} as a teacher.

\subsection{Detection and Segmentation}\label{sec:a-transfer}
For detection and segmentation tasks, we fine-tune Mask R-CNN \cite{mask_rcnn} on COCO \cite{mscoco}. The typical FPN \cite{fpn} backbone is adopted. ``SyncBN" \cite{peng2018megdet} is adopted in both backbone and FPN. We fine-tune all layers (including BN) end-to-end. The schedule is the default 2$\times$ in \cite{detectron2}. Specifically, all the models are trained for 180k iterations with a batch size 16. The learning rate starts from 0.02, and is divided by 10 at \mbox{[120k, 160k]} iterations. The short-edge of the training image is in [640, 800], while fixed at 800 during inference. We adopt the same training settings for both MobileNet v3 \cite{mobilenetv3} and ResNet \cite{resnet}. 

In \cref{tbl:resnet-det} of the main paper, we adopt the pretrained ResNet18 \cite{resnet} provided by torchvision\footnote{\href{https://pytorch.org/vision}{https://pytorch.org/vision}.} for the \emph{Supervised-IN.} entry, and utilize the open-source pretrained checkpoints provided by the authors for SEED \cite{SEED} and CompRess \cite{CompRess}. The teacher models for them are a 2$\times$ wider ResNet50 \cite{resnet} network pretrained for 400 epochs with SwAV \cite{swav} method and a ResNet50 network pretrained for 800 epochs with SwAV \cite{swav} method respectively. 

\subsection{Image Size}
During self-supervised pretraining, we resize the images to 128$\times$128 for ResNet \cite{resnet}, which is smaller than the common setting (224$\times$224). Here we provide an alternative setup, experimentally analyze the effects of different image size settings by testing ResNet18 with BYOL-based pretraining for 300 epochs. When the images are resized to 224$\times$224, the total mini-batch size is reduced to 1024 from 4096. The above settings along with training cost and top1 accuracy obtained by linear evaluation are shown in \cref{tbl:size}. Two image size settings achieve comparable performance with also comparable training cost. As larger batch size makes it easier to from multi-node distributed training with more machines, which contributes to shorter training time, we thus adopt 128$\times$128 image size as the default setup.
\begin{table}[h]
\begin{center}
\setlength{\tabcolsep}{3.5pt}
\small
\vspace{-10pt}
\begin{tabular}{cccc}
\textbf{Batch size} & \textbf{Image size} & \textbf{Training cost} (GPU hour) & \textbf{Top1 accuracy} \\
\toprule
1024 & 224 & 0.52k & 61.0 \\
4096 & 128 & 0.51k & 60.8 
\end{tabular}

\vspace{5pt}
\captionof{table}{\small Comparison of different training setups.}
\label{tbl:size}
\end{center}
\vspace{-35pt}
\end{table}

\section{DSPNet as the Initialization of Slimmable Networks}
\begin{figure}[ht]
    \centering
    \includegraphics[width=0.47\textwidth]{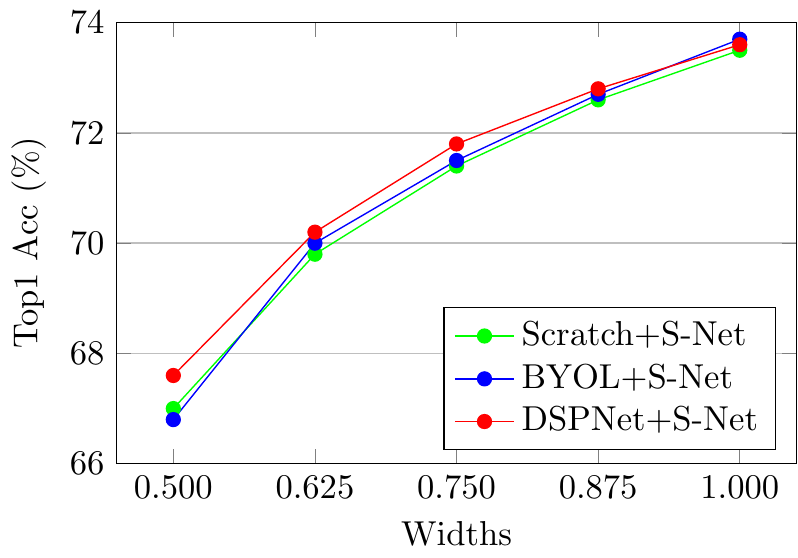}
    \caption{\small Top1 accuracy comparisons of slimmable networks \cite{slimmable} using different initialization methods. Our DSPNet can serve as an ideal pretraining.}
    \label{fig:r-slimmable}
\vspace{-20pt}
\end{figure}

In this section, we investigate the value of DSPNet on promoting the performance of slimmable networks \cite{slimmable}. Slimmable networks \cite{slimmable, slimmablev2, bignas, cai2019once} were introduced that can switch among different widths and depths at runtime, permitting instant and adaptive accuracy-efficiency trade-offs. In this sense, our DSPNet can serve as an ideal pretraining method for slimmable networks \cite{slimmable}. To validate this assumption, we conduct the DSPNet-based pretraining with MobileNet v3 at widths $[0.5, 0.625, 0.75, 0.875, 1.0]$, and then fine-tune the obtained full-size online encoder following the training algorithm of slimmable networks \cite{slimmable} on the ImageNet \cite{ImageNet} classification task. The switches of the slimmable network during the supervised training are the same as during our DSPNet-based pretraining, \ie, MobileNet v3 at widths $[0.5, 0.625, 0.75, 0.875, 1.0]$. The \emph{sandwich rule} \cite{slimmablev2} and \emph{inplace distillation} \cite{slimmablev2} training techniques are adopted. We adopt the training recipes provided by \texttt{timm} \cite{rw2019timm}, except that the number of training epochs is reduced to 300 from 600 for fast experiment. We show the effect of different initialization strategies for training slimmable networks and report the top1 accuracy of the five switches of them in \cref{fig:r-slimmable}. 

 In the comparison, we first adopt a pretrained MobileNet v3 at width $1\times$ with BYOL \cite{BYOL} as the initialization of the supervised training slimmable network (BYOL+S-Net in \cref{fig:r-slimmable}). However, no improvement is observed compared with that trained from the scratch (Scratch+S-Net in \cref{fig:r-slimmable}). Then we adopt our DSPNet as the initialization (DSPNet+S-Net in \cref{fig:r-slimmable}), and improvements are observed especially for the smaller switches. We attribute it to that DSPNet provides a more suitable initialization for the subsequent supervised training of slimmable networks, in which all the desired sub-networks have relatively good representations at the very beginning of the training.

\section{Training Stability}

\begin{figure}[ht]
    \centering
    \includegraphics[width=0.5\textwidth]{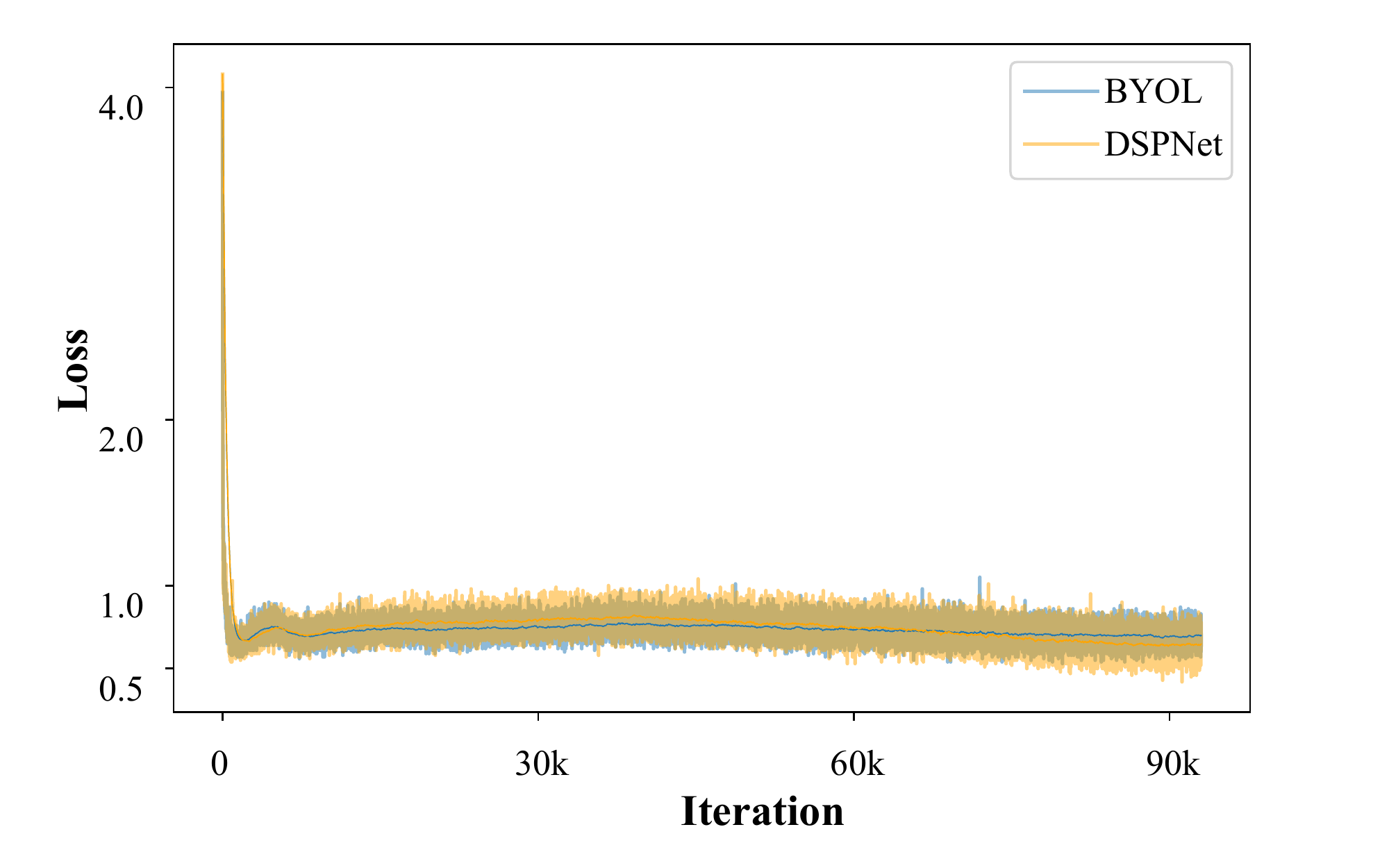}
    \vspace{-5pt}
    \caption{\small Training curves of BYOL and our DSPNet.}
    \label{fig:r-loss}
\vspace{-25pt}
\end{figure}

Since randomness is involved during the training of our DSPNet, the training stability should be considered. We provide the training curves in \cref{fig:r-loss}, which show only a little more aggravated training perturbation for our DSPNet than BYOL baseline. Furthermore, our 3 tests with different random seeds achieve similar performance. 
\vspace{-15pt}
\end{document}